\documentclass[runningheads]{llncs}
\usepackage{booktabs}
\usepackage{algorithm}
\usepackage{algorithmic}

\usepackage{hyphenat}
\usepackage{multirow}
\usepackage{xcolor}
\usepackage{times}
\usepackage{soul}
\usepackage{url}
\usepackage[hidelinks]{hyperref}

\usepackage[utf8]{inputenc}
\usepackage[normalem]{ulem}

\usepackage[hidelinks]{hyperref}
\usepackage[utf8]{inputenc}
\usepackage[small]{caption}
\usepackage{graphicx}
\usepackage{amsmath}
\usepackage{amssymb}
\usepackage{booktabs}
\usepackage{algorithm}
\usepackage{algorithmic}
\usepackage[switch]{lineno}
\usepackage{float}
\usepackage{subcaption}
\usepackage[normalem]{ulem}

\usepackage[T1]{fontenc}
\usepackage{graphicx}
\begin{document}
\title{Dynamic Label Adversarial Training for Deep Learning Robustness Against Adversarial Attacks}
\titlerunning{Dynamic Label Adversarial Training}
%
\author{
Zhenyu Liu\inst{1} \and
Haoran Duan\inst{1} \and
Huizhi Liang\inst{1} \and
Yang Long\inst{2} \and
Vaclav Snasel\inst{3}\and
Guiseppe Nicosia\inst{4}\and
Rajiv Ranjan\inst{1} \and
Varun Ojha\inst{1}
}
\authorrunning{Z. Liu et al.}

\institute{Newcastle University, Newcastle, UK \and
Durham University, UK \and
University of Catania, Catania, Italy \and
Technical University of Ostrava, Ostrava, Czech Republic
}

\maketitle              
\begin{abstract}

Adversarial training is one of the most effective methods for enhancing model robustness. Recent approaches incorporate adversarial distillation in adversarial training architectures.
However, we notice two scenarios of defense methods that limit their performance: (1) Previous methods primarily use static ground truth for adversarial training, but this often causes robust overfitting; (2) The loss functions are either Mean Squared Error or KL-divergence leading to a sub-optimal performance on clean accuracy. To solve those problems, we propose a dynamic label adversarial training (DYNAT) algorithm that enables the target model to gradually and dynamically gain robustness from the guide model’s decisions. Additionally, we found that a budgeted dimension of inner optimization for the target model may contribute to the trade-off between clean accuracy and robust accuracy. Therefore, we propose a novel inner optimization method to be incorporated into the adversarial training. This will enable the target model to adaptively search for adversarial examples based on dynamic labels from the guiding model, contributing to the robustness of the target model. Extensive experiments validate the superior performance of our approach.    

\end{abstract}

\section{Introduction}\label{sec:dynat_intro}
\label{sec:intro}
Deep neural networks (DNNs) have demonstrated remarkable success across various applications. Their widespread adoption has brought an important concern to the forefront. DNNs have been found susceptible to adversarial attacks that are crafted by imperceptible perturbations to inputs to fool DNNs~\cite{goodfellow2014explaining,szegedy2013intriguing,huang2023boosting}. Moreover, this weakness of DNNs makes many real-world applications such as image classification~\cite{goodfellow2014explaining}, image segmentation~\cite{goodfellow2014explaining}, vision transformers~\cite{zhang2024improving},  and others 
vulnerable to such adversarial attacks.

Several studies have addressed this problem by proposing \textit{adversarial training} of DNN models to improve their robustness against adversarial attacks (e.g.,~\cite{zhu2023towards,jia2022adversarial}). The adversarial training improves model robustness by incorporating adversarial examples into the training process through a \textit{minimax optimization} approach. Such a form of adversarial defense against adversarial attacks has become an ongoing essential research topic in computer vision. The adversarial training~\cite{madry2017towards} is regarded as one of the most effective defense methods to secure DNNs. Under adversarial training, the class label plays a key role in the loss function.
\begin{figure}[t]
\centering
\includegraphics[width=1.00\linewidth]{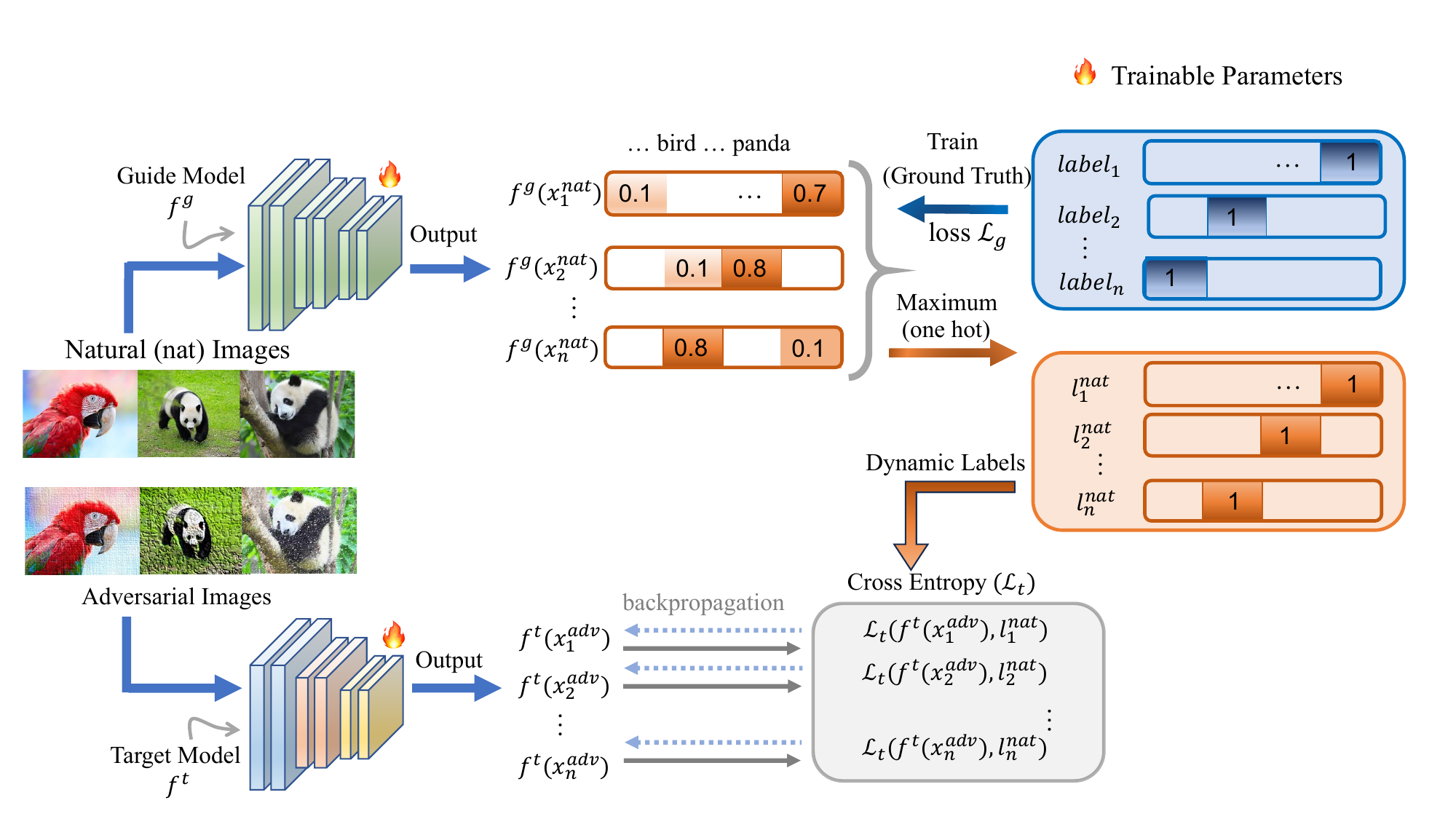}
\caption{Proposed dynamic label adversarial training (DYNAT) of deep learning models. DYNAT explicitly gives flexibility on the loss functions for adversarial training, whose dynamic label comes from the guiding model. We train the guide model/network using a dataset with static (ground truth) labels (blue labels, $label_1, label_2, \ldots$.) and concurrently use the maximum value in the guiding model's logits as \textit{dynamic labels} (orange labels, $l_1^{nat}, l_2^{nat}, \ldots$) for adversarial training of the target model/network. The guiding model $f^g (x_i^{nat})$ takes a clean image \(x_i^{nat}\) as its input and produces a softmax probability value vector, which is converted into labels $l_i^{nat}$ using a one-hot or winner-takes-all principle that is used by target model for computing its cross-entropy $\mathcal{L}_t(f^t (x_i^{adv}),l_i^{nat})$ between target model's $f^t (x_i^{adv})$ on adversarial image \(x_i^{adv}\). (The adversarial example generated in each iteration of training using an inner optimization is shown in Fig.~\ref{fig:inner_optimise}.) This cross-entropy loss on dynamic label backpropagates to the target model for its dynamic label adversarial training. The dynamic label strength increases from weak to strong as the training loss $\mathcal{L}_t$ (Eq.~\ref{eq:teacher_student_loss}) minimizes iteratively.}
\label{fig:DYNAT}
\end{figure}

Moreover, the recent adversarial training advancements incorporate \textit{adversarial distillation} training architectures, which are among the most successful adversarial training methods~\cite{goldblum2020adversarially}. Adversarial distillation enhances the adversarial robustness of DNNs with a teacher-student framework. Like adversarial training, adversarial distillation can be defined as a minimax optimization problem that aims to transfer both prediction accuracy and adversarial robustness from a larger robust teacher model to a student model. Methodologically, existing adversarial distillation methods proposed in~\cite{goldblum2020adversarially,zi2021revisiting,huang2023boosting} employ \textit{soft labels} (i.e., the softmax outputs) generated by a teacher DNN model to guide the training of a student DNN model. 
Typically, these adversarial distillation scenarios employ Kullback-Leibler (KL) divergence as loss functions to approximate probability distributions between the teacher and student models. However, we observe that the methodologies employed in traditional adversarial training and adversarial distillation constrain the clean performance of the target model due to: (1) DNNs are trained using static labels (ground truth) in existing standard adversarial training methods~\cite{madry2017towards}. (2) In the current teacher-student framework, the loss functions are either Mean Squared Error (MSE)~\cite{cui2021learnable} or KL-divergence that guides the student model. These applications of KL-divergence and ground truth in defense methods have improved the robustness of the model, but its performance on clean accuracy remained sub-optimal.

We hypothesize that the adversarial training process can be efficient (a) if it uses dynamic labels (i.e., iteratively generated classification outputs/labels) instead of using static ground truth labels and (b) if it uses cross-entropy loss function on dynamically generated labels from a clean guiding model, instead of using a pre-trained teacher model with KL divergence because the cross-entropy function directly focuses on the classification capabilities of models. 


Therefore, we propose a dynamic label adversarial training (DYNAT) algorithm that enables the target model to gradually and dynamically gain robustness from the guide model's decisions.  We further explain the benefits of the guide model in Sec.~\ref{sec:learning_dynamic_AT}. This effectively mitigates the over-fitting issue~\cite{chen2020robust,rice2020overfitting}, thereby enhancing the target model's clean accuracy (accuracy on natural images) and robust accuracy (accuracy on adversarial images).  

In addition to the DYNAT algorithm, we also found that a budgeted dimension of \textit{inner optimization}~\cite{huang2023boosting} (i.e., the budget of adversarial example generation) for the target model may also contribute to the trade-off between clean accuracy and robust accuracy. To ease the budget of adversarial example generation and over-fitting phenomenon, we propose a novel \textit{inner optimization} method to be incorporated into the adversarial training. This will enable the target model to adaptively search for adversarial examples based on dynamic labels from the guiding model, contributing to the robustness of the target model. Hence, the contribution of this paper is three-fold.
\begin{itemize}
\item We propose a simple yet effective novel \textit{dynamic label adversarial training} (DYNAT) algorithm. This is based on a novel mix of adversarial distillation and adversarial training. The DYNAT algorithm performs adversarial training in a ``weak to strong'' fashion, effectively alleviating the issue of overfitting phenomena.

\item Our DYNAL algorithm is supplemented by a novel \textit{dynamic inner optimization} method for adversarial examples generation. This aims to gradually and adaptively generate adversarial examples for the adversarial training of DNNs. 

\item Compared to existing defense methods, our DYNAT algorithm yields superior and competitive performance on a varied state-of-the-art adversarial attack test bench. This is when we validate the effectiveness of our DYNAT algorithm on an extensive set of experiments using CIFAR-10~\cite{krizhevsky2009learning} and CIFAR-100~\cite{krizhevsky2009learning} datasets with two model architectures ResNet~\cite{he2016deep} and WideResNet~\cite{zagoruyko2016wide} that were alternatively used as guiding and target models in our experiment settings. 
\end{itemize}

\section{Background and Related Work}\label{sec:dynat_realted_work}
\label{sec:related_work}
Our work on adversarial defense is related to research in adversarial training, adversarial distillation, and adversarial defense. We discuss each of these topics as follows:

\noindent \textbf{Adversarial Attacks.}  Given the acknowledged vulnerabilities in deep learning models~\cite{pravin2021adversarial}, some studies have proposed a series of adversarial attack methods. Adversarial examples were initially proposed by Szegedy et al.~\cite{szegedy2013intriguing,goodfellow2014explaining}, and the classic method for generating them, called the Fast Gradient Sign Method (FGSM), proved to be a simple and effective approach. Madry et al.~\cite{madry2017towards} later proposed a Projected Gradient Descent (PGD) attack method, an enhanced version of FGSM, employing multiple iterations to generate adversarial examples. Similarly,  AutoAttack (AA)~\cite{andriushchenko2020square,croce2020minimally}, including its two variant methods (APGDCE and APGD-DLR) of the PGD-attack were developed as effective DNNs attack methods. DeepFool~\cite{moosavi2016deepfool} was designed as an iterative attack method that calculates perturbations by linearly approximating the gradient of the target function. Papernot et al.~\cite{papernot2016limitations} proposed Jacobian-based Saliency Map Attacks (JSMA). Subsequently, Carlini and Wagner~\cite{carlini2017towards} proposed the C\&W attack, offering a solution to an optimization problem for minimizing perturbations added to the original input to attack DNN models. Several defense strategies (as discussed below) were proposed to mitigate these attacks. 

\noindent \textbf{Adversarial Defense.} As a defense scheme, adversarial training leverages worst-case optimization to enhance model robustness, encompassing approaches such as adversarial training and its variants. The method TRADES by Zhang et al.~\cite{zhang2019theoretically} has been explored as a defense method to trade-off between the clean accuracy and robust accuracy of the target model. Some research indicates that, under limited model capacity, incorporating weight information contributes to improved model robustness. Thus, Wu et al.~\cite{wu2020adversarial} investigated the adversarial weight perturbation (AWP) robustness method by adding perturbations to model weights. Additionally, Cui et al.~\cite{cui2021learnable} proposed learnable boundary guidance adversarial training (LBGAT), using logits generated by a guiding model to guide the learning process of a target model. Similarly, geometry-aware instance-reweighted adversarial training (GAIRAT)~\cite{zhang2020geometry} has achieved excellent performance on PGD attacks. However, the margin-aware instance re-weighting learning (MAIL) method~\cite{liu2021probabilistic} highlighted the weakness of the GAIRAT method and improved its performance on C\&W and AA attacks. These defense methods are based on the teacher-student framework that performs adversarial distillation. 

\noindent \textbf{Adversarial Distillation.} Large-capacity models showed better performance in adversarial training than smaller-capacity models~\cite{schmidt2018adversarially}. Hinton et al.~\cite{hinton2015distilling} proposed adversarial distillation to enhance the effectiveness of adversarial training for smaller models. This method aims to transfer knowledge from a larger, robust model (teacher model) to a smaller model, which is more suitable for low-cost computation, thereby improving the robustness of the smaller model (student model). This has been extensively researched and proven to be promising. However, LBGAT~\cite{cui2021learnable} shows that smaller guidance models can improve target models' cleaning accuracy performance. From the perspective of knowledge distillation, we explore that LBGAT differs from adversarial distillation in two aspects: (a) the guiding model is only trained on clean samples without distillation, and (b) the target model in LBGAT does not inherit the robustness of the guiding model. LBGAT shows impressive performance on clean accuracy. Hence, we adopt these two properties in our DYNAT method.  Notably, the models are typically pre-trained in most adversarial knowledge distillation work. However, similar to LBGAT, we emphasize simultaneous training of the guiding and target models from the outset in our work. 

As opposed to traditional adversarial training and adversarial distillation methods, where teacher and student models are trained on static labels (using ground truth) only, this work seeks to understand how the target model inherits the implicit properties of the guiding model without directly relying on the guiding model's logits and ground truth labels.

\section{Methodology}\label{sec:DYNAT}
We introduce a novel guide framework by incorporating the ``weak-to-strong'' concept for adversarial training. In Sec.~\ref{sec:priliminary}, we provide a preliminary on adversarial training. Subsequently, in Sec.~\ref{sec:DYNAT}, we present our novel DYNAT algorithm, and in Sec.~\ref{sec:learning_dynamic_AT}, we analyze the learning processes of DYNAT. In Sec.~\ref{sec:inner_opt}, we propose a novel inner optimization for adversarial example generation. We analyze the convergence of DYNAT in Sec.~\ref{sec:convergance}.

\subsection{Preliminary}\label{sec:priliminary}
\paragraph{Adversarial Training Process.} 
We first review adversarial training in the ordinary setting. Let the training set of $n$ instance-label pairs be $D(X, Y) = \lbrace (x_i, y_i) \rbrace_{i = 1}^n$ and $x_i \in \mathbb{R}^d$ be a clean example, $Y \in \lbrace 1, \ldots, C \rbrace$ be a set of $C$ class labels such that $y_i \in Y$ is assigned to an instance $x_i \in  X$. Additionally, $\mathbb{L}$ defines the set of allowed perturbations on a clean example $x_i$. The perturbation is usually constrained by $L_p$ norm with a bound $\epsilon$, 
\begin{align}
    \label{eq:perturbation}
    \mathbb{L}=\lbrace \delta ~|~ \lVert \delta \rVert_p \leq \epsilon \rbrace.
\end{align}

In this setting, given \ target image $x$, adversarial training aims to find a perturbation $\delta \in \mathbb{R}^d$, on the input from the target model through a minimax process. The standard adversarial training is formulated as a minimax optimization problem as follows:
\begin{equation}
\label{eq:adv_training}
\mathop{\text{minimize}}_{\theta} \frac{1}{|D|} \sum_{x,y \in D} \mathop{\max}_{\delta\in \mathbb{L}}  \mathcal{L}(\theta, f(x + \delta), y),
\end{equation}
where $\theta$ is a model's learnable parameters, $f(x + \delta)$ is model's output of attacked input, and $\mathcal{L}$ is a cross-entropy loss function. 

\subsection{Dynamic Label Adversarial Training (DYNAT) Method} \label{sec:DYNAT}
In this section, we present our DYNAT algorithm. Unlike standard adversarial training, we do not adopt ground truth as training labels. Instead, our goal is to use dynamic labels. These dynamic labels come from the guide model and are defined as: 
\begin{align}
    \label{eq:dynamic_lable}
    l^{nat}  = \text{onehot} \{f_{g}(x) \},
\end{align}
where $x$ is a clean image, $f_{g}(\cdot)$ represents the guiding model output, and $l^{nat}$ is the one hot, i.e., winner label of guiding model's logits (SoftMax outputs; see Fig.~\ref{fig:DYNAT}). As shown in Fig.~\ref{fig:DYNAT}, the labels for adversarial training evolve dynamically from ``weak'' (classification capabilities of the guiding model in initial iterations) to ``strong'' (classification capabilities of the guiding model as it approaches convergence). For the extraction of dynamic labels, we select $l^{nat}$, which is the maximum logit in the guiding model’s output as the \textit{dynamic label}. This dynamic label follows a weak to strong rhythm, i.e., as the training progresses, we hypothesize the dynamic label will become stronger (accurate in classifying clean examples). The dynamic label is continuously obtained via clean training the guiding model by minimizing the following clean loss:
\begin{align}
\label{eq:teacher_loss}
    \mathop{\text{minimize}}_{\theta_g} \frac{1}{|D|} \sum_{x,y \in D} \mathrm{ \mathcal{L}}_g (\theta_g, f_{g}(x), y),
\end{align} 
where $\mathcal{L}_g$ is guiding  model loss, $f_{g}(x)$ is output of guiding model on clean input $x$,  $\theta_g$ is the guiding model's parameter, and $y$ is ground truth. We employ the cross-entropy function $\mathcal{L}_g$ for clean classification training of the guide model, as it consistently yields fast and highly accurate clean outputs, surpassing the clean accuracy of the target model in adversarial training.

We minimize the adversarial loss of the target model (optimize the target model parameter $\theta_t$) using dynamic labels $l^{nat}$ coming from the guiding model in each iteration of training as:
\begin{align}
\label{eq:student_loss}
    \mathop{\text{minimize}}_{\theta_t} \frac{1}{|D|} \sum_{x,y \in D} \mathrm{ \mathcal{L}}_t (\theta_t, f_{t}(x+ \delta), l^{nat}) ,
\end{align}
where $f_{t}(x + \delta)$ are outputs of the target model on adversarial examples, and $\mathcal{L}_t$ is the loss between the target model's output and dynamic label from the guiding model. In other words, $\mathcal{L}_t$ is the loss for target model adversarial training.

As in Eq.~\eqref{eq:adv_training}, adversarial training adopts the cross-entropy function in classification tasks. However, the use of static labels leads to robust overfitting~\cite{chen2020robust}~\cite{rice2020overfitting}, resulting in poor generalization to clean samples. Our approach aims to mitigate issues of robust overfitting by using dynamic labels via a guide model while using the cross entropy function\footnote{Some other methods such as LBGAT~\cite{cui2021learnable} uses Mean Square Error and some other employ Kullback-Leibler divergence.} as the $ \mathcal{L}_t$ loss function for adversarial training. 
See the orange color arrow in  Fig.~\ref{fig:DYNAT} indicating the flow of dynamic label into the calculation of $\mathcal{L}_t$ loss. In this loss function, an instance $x + \delta$ is associated with the maximum adversarial perturbation and is fed as an input to the target model, and $l^{nat}$ is fed as a training (dynamic) label instead of a fixed ground truth from the dataset. Overall, we train both guiding and target models simultaneously by balancing their losses as follows:
\begin{align}
\label{eq:teacher_student_loss}
    \mathop{\text{minimize}}_{\theta_g, \theta_t} \frac{1}{|D|} \sum_{x,y \in D} \big\{\mathrm{ \mathcal{L}}_g(\theta_g, f_{g}(x),y) + \beta  \mathrm{ \mathcal{L}}_t(\theta_t, f_{t}(x + \delta),l^{nat})\big\},
\end{align}
where $\beta \in [0,1]$ is a user defined hyperparameter. 
We expect the target model to dynamically follow the guiding model-generated prediction labels during the training process. From Eq.~\eqref{eq:student_loss}, it can be observed that the target model follows the guiding model in the minimization process of the two models' architecture, which is also summarized in Algorithm~\ref{alg:1}.
\begin{algorithm}[t]
   \caption{Dynamic label adversarial training method}
   \label{alg:1}
\begin{algorithmic}
   \STATE {\bfseries Input:} Guiding network/model $f_g$, target network/model $f_t$, training data $D$, mini-batch  $X$, batch size $m$,  $\beta$ is a hyper-parameter.
   \STATE {\bfseries Output:} Adversarially robust model $\theta_t$.
   \REPEAT
   \STATE Read mini-batch $X$ from training set $D$.
   \STATE Get adversarial examples $X^{adv} = \{x_{1}^{adv}, ..., x_{m}^{adv} \}$ by adversarial attack on input $X$
   \STATE Generate $ logits^{guide} = f_g (X)$ from guiding model
   \STATE Generate $ logits^{target} = f_t (X_{adv})$ from target model
   \STATE Convert $logits^{guide}$  to dynamic label $l^{nat}$ as per Eq.~\eqref{eq:dynamic_lable}
   \STATE Compute guiding model loss $\mathcal{L}_g$ as per Eq.~\eqref{eq:teacher_loss}
   \STATE Compute target model loss $\mathcal{L}_t$ as per Eq.~\eqref{eq:student_loss}
   \STATE Optimize parameters of both guiding and target models $\theta_t$ by minimizing the loss in Eq.\eqref{eq:teacher_student_loss}
   \UNTIL{training converged}
\end{algorithmic}
\end{algorithm}

\subsection{Analysis of DYNAT} \label{sec:learning_dynamic_AT}

In this section, we analyze the learning process of the DYNAT method shown in Fig.~\ref{fig:DYNAT}. Our analysis will cover the following three aspects: the learning process, comparison with previous methods, and the benefits of dynamic labels.

\noindent \textbf{Learning process.}  In DYNAT, the learning of the target model is guided by the guiding model's outputs (dynamic labels) that gradually come close to ground truth labels (see orange labels in Fig.~\ref{fig:DYNAT}). At the beginning (earlier epochs of training) of adversarial training, the guiding model's classification performance will be suboptimal, leading to weak adversarial training (i.e., weak dynamic labels generated by the guiding model) in the target model. Notably, the guide model consistently maintains superior classification accuracy compared to the target model, leading to stronger adversarial training (i.e., strong, robust, and closer to ground truth dynamic labels generated by the guiding model) into the target model. These dynamic labels approximate ground truth labels and guide the target model during the adversarial training. We call this method of dynamic learning a ``\textit{weak-to-strong}'' adversarial training method.    

\noindent \textbf{Comparison with previous method:} Compared to the typical adversarial distillation method, our method of adversarial training is inherently different because of the dynamic label use. In other words, a typical adversarial distillation method aims to distill the robust knowledge of a larger, well-trained teacher model into small student models. Unlike this, \textit{our target model does not necessarily need to inherit the robustness from a large model} to achieve satisfactory robust performance as any size of well-performing guiding model would generate strong dynamic labels. 

\noindent \textbf{Salient points of DYNAT.}
Our method aims to make the labels dynamic without emphasizing the differences between the guiding model and target model probability distributions. Hence, similar to one of the most effective defense methods, adversarial training, the target model in training will directly and effectively focus on the classification task rather than the probability distribution. 

It is worth noting the benefits of using the guiding model: (i) the guiding model maintains a higher clean accuracy compared to the target model because the generalization capability of the target model on clean samples is lower than that of the guiding model trained directly on clean samples. (ii) despite the possibility of the guiding model generating incorrect dynamic labels, the guiding model still consistently maintains a higher clean accuracy than the target model, providing guidance for the target model to gain robustness. 

\subsection{Inner Optimization: Dynamic Adversarial Examples Generation}\label{sec:inner_opt}
In most previous related works, the adversarial examples are generated via a single DNN model. For example, this is done by using cross-entropy on the softmax output (logit) of a model and the ground truth~\cite{rice2020overfitting} or by using KL divergence between clean logits and the adversarial logits~\cite{zhang2019theoretically}. On the other hand, authors in~\cite{chen2020robust} found that the over-fitting phenomenon exists in adversarial training. This ``overfitting'' issue arises when the target model is trained on adversarial examples generated with static ground truths, potentially impacting the model's generalization to defend against other types of attacks and classify clean samples.

\begin{figure}[h!]
\centering
\includegraphics[width=1.\linewidth]{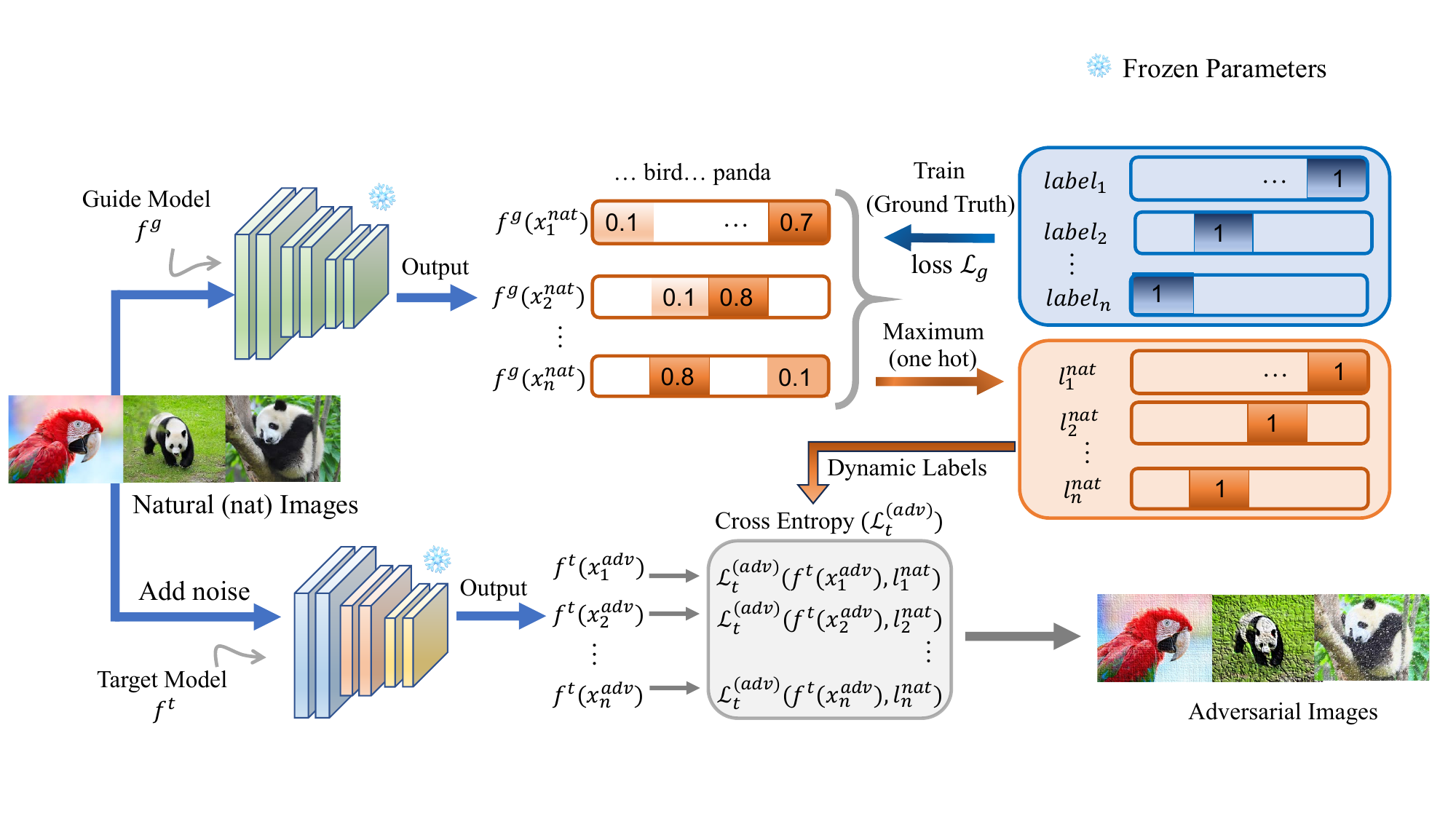}
\caption{Our inner optimization framework. The natural images are fed into fixed target and guiding models. Then, we use our strategy to extract dynamic labels from guiding model outputs (orange labels). We encourage target model outputs and dynamic labels to participate together in adversarial example generation via the cross-entropy function. This generated adversarial example is fed back to the target model for the dynamic label adversarial training, i.e., outer optimization as shown in Fig.~\ref{fig:DYNAT}. That is, Fig.~\ref{fig:DYNAT} and Fig.~\ref{fig:inner_optimise} are snapshots of the same training iteration where the target model (see the snow symbol in Fig.~\ref{fig:inner_optimise} indicating the target model's parameters are frozen) in Fig.~\ref{fig:inner_optimise} first produced adversarial example based on the dynamic label from the guiding model and Fig.~\ref{fig:DYNAT} takes this adversarial example in the same training iteration to perform adversarial training (see the fire symbol in Fig.~\ref{fig:DYNAT} indicating update of target model's parameters).}
\label{fig:inner_optimise}
\end{figure}

We propose a novel inner optimization (dynamic adversarial example generation method) aiming at searching for the ``support'' adversarial example $x + \delta $ that maximizes the prediction differences between the target model and the guide model under dynamically increasing label strength (see Fig.~\ref{fig:inner_optimise}).
Our method employs the PGD~\cite{rice2020overfitting} attack to generate adversarial examples for adversarial training. However, our method differs from the PGD-AT~\cite{rice2020overfitting} that uses static labels for its adversarial example generation. The adversarial examples generated by our method (shown in Fig.~\ref{fig:inner_optimise}) are fed to the DYNAT learning process (shown in Fig.~\ref{fig:DYNAT} and described in Sec.~\ref{sec:DYNAT}). We obtain the optimal maximum norm-constrained adversarial example $x^{adv}$ and the dynamic label around the fixed parameter $\theta_t $ using the cross-entropy loss $\mathcal{L}_t^{(adv)}$ as:
\begin{equation}
\label{eq:inner_opt}
    x^{adv} = x +  \mathop{arg \max}_{\delta\in \mathbb{L}}  \mathcal{L}_t^{(adv)}(\theta_t, f_t(x+\delta), l^{nat}).
\end{equation}

Although we observed that the guiding model exhibits initial weak predictions in the early stages of training, resulting in lower output accuracy, it maintains high classification performance on clean images in the later stages of training and consistently surpasses the target model in clean accuracy. Consequently, the target model can progressively learn to generate stronger adversarial examples by focusing on highly accurate dynamic labels. 
Inner optimization allows the target model to dynamically learn adversarial examples from weak to strong to improve the robustness of the model. This ``weak to strong'' adversarial example generation may mitigate the over-fitting phenomenon. Unlike TRADES~\cite{zhang2019theoretically}, which is a single-model adversarial training method that uses logits with KL divergence to guide their inner optimization, our method uses \textit{dynamic label}s based on guidance model and cross-entropy to guide the inner optimization.

\subsection{Convergence Analysis} \label{sec:convergance}
\begin{figure}[t]
\begin{subfigure}[t]{0.49\linewidth}
        \centering
        \includegraphics[width=1\linewidth]{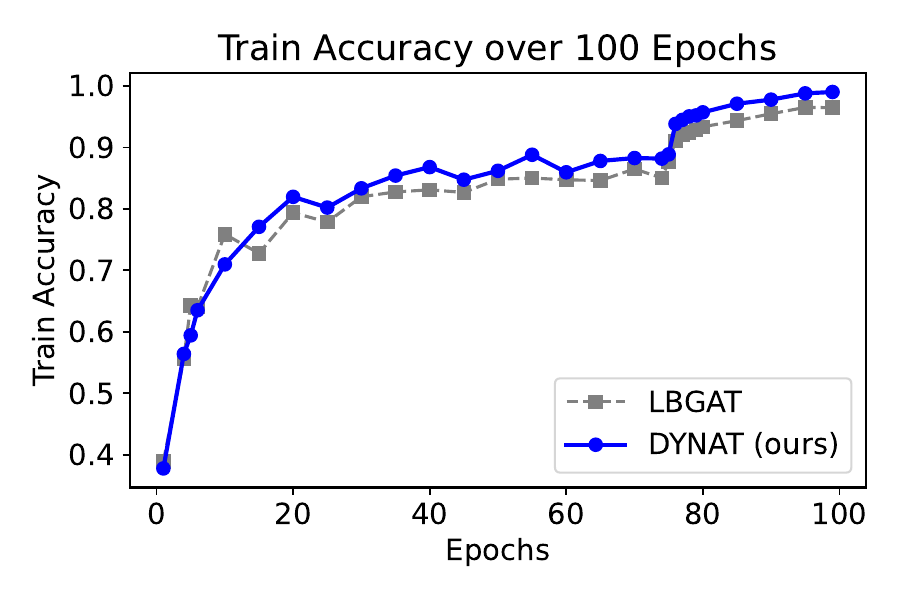}
        \caption{}
    \end{subfigure} \hfill
\begin{subfigure}[t]{0.49\linewidth} 
        \centering
        \includegraphics[width=.98\linewidth]{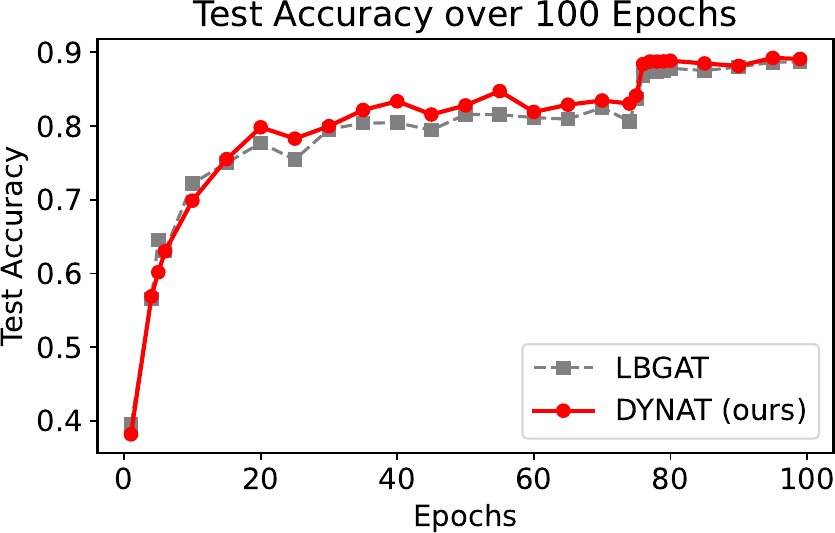}
        \caption{}
    \end{subfigure} 
  \caption{Epoch-wise performance of DYNAT compared with LBGAT on CIFAR-10 dataset. The student model/network was trained on 100 epochs. (a) Train accuracy of the target model and (b) Test accuracy of the target model.  Note that gray lines represent the LBGAT method, and blue/red lines represent our proposed DYNAT method. 
  }
\label{fig:convergance}
\end{figure}

We performed an empirical convergence analysis of DYNAT and compared its performance against LBGAT to understand the effect of dynamic labels on clean classification accuracy. We observe that the target model shows the same convergence trajectory as  LBGAT on both training and test sets. However, DYNAT demonstrates superior accuracy overall, confirming the training status of the target model from the guidance model in the form of dynamic labels (see Fig.~\ref{fig:convergance}). 

\section{Experiments and Results} 
Our experiments consistently yielded identical results, with both final outcomes and loss values remaining constant across multiple runs under the same parameters. This demonstrates the reproducibility and determinism of our experimental setup, ruling out randomness.
\label{sec:experiments}
\subsection{Experiment Set up}
\textbf{DYNAT and its Variants.} We conduct a series of expansive and intensive experiments to validate the effectiveness of our proposed DYNAT method and its variants: DYNAT-AWP (DYNAT with AWP~\cite{wu2020adversarial}); DYNAT-Inner-AWP (DYNAT with Inner optimization and AWP); \textit{DYNAT-GAIRAT} (DYNAT with GAIRAT~\cite{zhang2020geometry}); DYNAT-Inner-GAIRAT (DYNAT with Inner optimization and GAIRAT); DYNAT-MAIL (DYNAT with MAIL \cite{liu2021probabilistic}); and DYNAT-Inner-MAIL (DYNAT with Inner optimization and MAIL).  

\noindent \textbf{Datasets.} We evaluate these methods on CIFAR-10 (with 10 classes) and CIFAR-100 (with 100 classes) datasets, each containing 60,000 images of pixel size 32×32 and having 50,000 training images and 10,000 test images.

\noindent \textbf{Test Attack Methods.} We select five state-of-the-art attack methods in increasing order of attack strength to \textit{test} the robustness of our methods: PGD-10, PGD-20, PGD-50, C\&W, AA.





\noindent \textbf{Training Attack Method.} In our experiment, the adversarial training method used adversarial examples generated by PGD to train the model to enhance its robustness. We follow the default setting of LBGAT ~\cite{cui2021learnable}; the perturbation is set to 0.031.  The step sizes of PGD are 2/255, respectively. We set the batch size to 128.

\noindent\textbf{Baseline.} We compare DYNAT and its variants with the following adversarial training methods: PGD-AT~\cite{rice2020overfitting}, 
TRADES ~\cite{zhang2019theoretically},
SAT~\cite{sitawarin2021sat},
MART~\cite{wang2019improving},
FAT~\cite{zhang2020attacks},
GAIRAT~\cite{zhang2020geometry},
AWP~\cite{wu2020adversarial} and
LBGAT~\cite{cui2021learnable}. To ensure the efficacy of dynamic labels, we also combined DYNAT with GAIRAT and MAIL and compared the results. Our experiment is based on PyTorch and the code that is publicly available\footnote{https://github.com/lusti-Yu/Dynamic-label.git}.

\noindent \textbf{DYNAT Loss Trade-off.} 
The hyper-parameter $\beta$ controls the alternative update between two loss functions as mentioned in Eq.~\eqref{eq:teacher_student_loss}. On CIFAR-10, we set $\beta$ as 1, giving equal weight to both loss functions. In CIFAR-100, when training DYNAT, we set $\beta$ to 1, but when training its variants DYNAT-AWP and DYNAT-Inner-AWP, we experimentally found $\beta =  0.2$  as the most suitable value.


\subsection{Comparisons among methods with WRN34-10 Architecture}
In the first set of experiments, we analyzed the effectiveness of \textit{our} methods and other methods where all methods use WideResNet34-10 (WRN34-10)~\cite{zagoruyko2016wide} network architecture as the target models. Our guiding model uses ResNet18~\cite{he2016deep} network architecture. The results of this comparison on CIFAR-10 and CIFAR-100 datasets are shown in Table~\ref{tb:cifar10_wrn34} and Table~\ref{tb:cifar100_wrn34}. 

On the CIFAR-10 dataset (see Table~\ref{tb:cifar10_wrn34}), our method achieves 89.16$\%$ in clean classification, with accuracy outperforming other baselines. Further, a variant of our method, DYNAT-AWP, improves the cleaning accuracy by 1.16\%, and the AA attack accuracy increases by 2.90\%. These results are better than the clean performance of other methods. Moreover, when our method is combined with AWP, inner, and TRADES (with its parameters $\alpha$ set to 1), our robustness on AA reaches 55.30$\%$, which outperforms all other methods. Our methods are highly competitive on PGD-10 (comes second) and marginally third on PGD-20 and PGD-50 other attacks. Our best-performing method, DYNAT-Inner-AWP($\alpha=1$), offers high competition across all attacks. In contrast, other best-performing defense methods are highly close to each other, and, on average, on all attacks, they perform best among other defense methods. 

From Table~\ref{tb:cifar100_wrn34}, our method improves the clean accuracy to 67.25\%. Moreover, our method (DYNAT-Inner-AWP, which is a combination of DYNAT with our Inner optimization and AWP and TRADES) achieves the best robustness performance under all attack scenarios. Specifically, our DYNAT-Inner-AWP method outperforms the other defense methods, such as AWP, by 0.98\% and 0.84\%  on the CW and AA attacks and LBGAT by 0.37\% on the AA attack. 

\begin{table}[t!]
\centering
\setlength{\tabcolsep}{2pt}
\caption{Test robustness accuracy (in \% ) on the CIFAR-10 dataset on WRN34-10 target model. $\alpha$ indicate a hyperparameter as in~\cite{zhang2019theoretically}, Inner indicates inner optimization as in Sec.~\ref{sec:inner_opt}, 'All' indicate average over all attacks, and \textbf{bold} and \underline{underlined} number indicates 1st and 2nd best.}
\label{tb:cifar10_wrn34}
\begin{tabular}{ll|ccccccc}
\toprule
& Method           & Clean          & PGD-10         & PGD-20         & PGD-50         & C\&W           & ~~AA~~  & ~~All           \\ \midrule 
\multirow{7}{*}{Others} 
& PGD-AT ~\cite{rice2020overfitting}         & 85.17          & 56.07          & 55.08          & 54.88          & 53.91          & 51.69 & 54.32 \\ 
& TRADES ~\cite{zhang2019theoretically}          & 85.72          & 56.75          & 56.1           & 55.9           & 53.87          & 53.40    & 55.21      \\ 
& MART ~\cite{wang2019improving}            & 84.17          & 58.98            &\underline {58.56}            &\underline {58.06}          & 54.58          & 51.10   & 56.25      \\ 
& FAT  ~\cite{zhang2020attacks}            & 87.97     & 50.31          & 49.86          & 48.79          & 48.65          & 47.48  & 49.02         \\ 
& GAIRAT ~\cite{zhang2020geometry}           & 86.30          & \textbf{60.64}          & \textbf{59.54}          & \textbf{58.74}          & 45.57          & 40.30  & 52.95  \\ 
& AWP  ~\cite{wu2020adversarial}            & 85.57          & 58.92          & 58.13          & 57.92            &\underline {56.03}          & 53.90   & \underline {56.98} \\
& LBGAT~\cite{cui2021learnable}           & {88.22}          &          56.25      &        54.66        &           54.3     &          54.29      & 52.20   & 54.34  \\ 

 \midrule 
\multirow{4}{*}{Ours} 
& DYNAT   &\underline {89.16}          &          52.88      &        51.78        & 51.49            &          50.88      & 49.20   & 51.25 \\ 

& DYNAT-AWP($\alpha$ = 0) & \textbf{90.27}         &          55.49       &        54.49        & 54.30           & 53.28            & 52.10  & 53.93  \\ 

& DYNAT-AWP($\alpha$ = 1) & 88.41         &          58.39  &        57.71        & 57.53           & 55.48           &\underline {54.10}   & 56.64  \\ 
& DYNAT-Inner-AWP($\alpha$ = 1) &  86.21     &\underline {59.08}  &    58.22         & 58.02            &\textbf{56.24}            & \textbf{55.30}  & \textbf{57.37}  \\ 
\bottomrule
\end{tabular}
\end{table}

\begin{table}
\centering
\setlength{\tabcolsep}{2.5pt}
\caption{Test robustness accuracy (\%)  on the CIFAR-100 dataset on WRN34-10 target model. $\alpha$ indicate a hyperparameter as in~\cite{zhang2019theoretically}, Inner indicates inner optimization as in Sec.~\ref{sec:inner_opt}, and \textbf{bold} and \underline{underlined} number indicates 1st and 2nd best results.}
 \label{tb:cifar100_wrn34}
\begin{tabular}{ll|cccccc}
\toprule
& Method           & Clean          & PGD-10         & PGD-20         & PGD-50         & C\&W             & AA             \\ \midrule 
\multirow{5}{*}{Others~} 
& PGD-AT~\cite{rice2020overfitting}           & 60.89          & 32.19          & 31.69          & 31.45          & 30.1           & 27.86          \\ 

& TRADES~\cite{zhang2019theoretically}           & 58.61          & 29.20          & 28.66          & 28.56          & 27.05          & 25.94          \\ 
& SAT~\cite{sitawarin2021sat}              &  \underline{62.82}          &     28.1           &       27.17         &  26.76              &    27.32            & 24.57          \\ 
& AWP~\cite{wu2020adversarial}              & 60.38          & 34.13          & 33.86          & 33.65          & 31.12          & 28.86          \\ 
& LBGAT~\cite{cui2021learnable}            & 60.64          &      35.13          &        \underline {34.75}        &             34.62   &     30.65           & 29.33          \\ \midrule 

\multirow{3}{*}{Ours~} & DYNAT  &  \textbf{67.25}  &     28.03         &      26.97         &      26.81         &      26.62        &       24.10   \\ 

& DYNAT-AWP ($\alpha$ = 1)  &{62.29}    & \underline{35.45}   & \textbf{35.09}  & \underline{34.92}  & \underline{31.50}  & \textbf{30.20}  \\ 

& DYNAT-Inner-AWP ($\alpha$ = 1)           &  58.87  &     \textbf{35.61}         &      \textbf{35.09}         &   \textbf{35.05}            & \textbf{32.10}             & \underline{29.70}       
 \\ \bottomrule
\end{tabular}
\end{table}

\begin{figure}[t]
\begin{subfigure}[t]{0.47\linewidth} 
        \centering
        \includegraphics[width=1.1\linewidth]{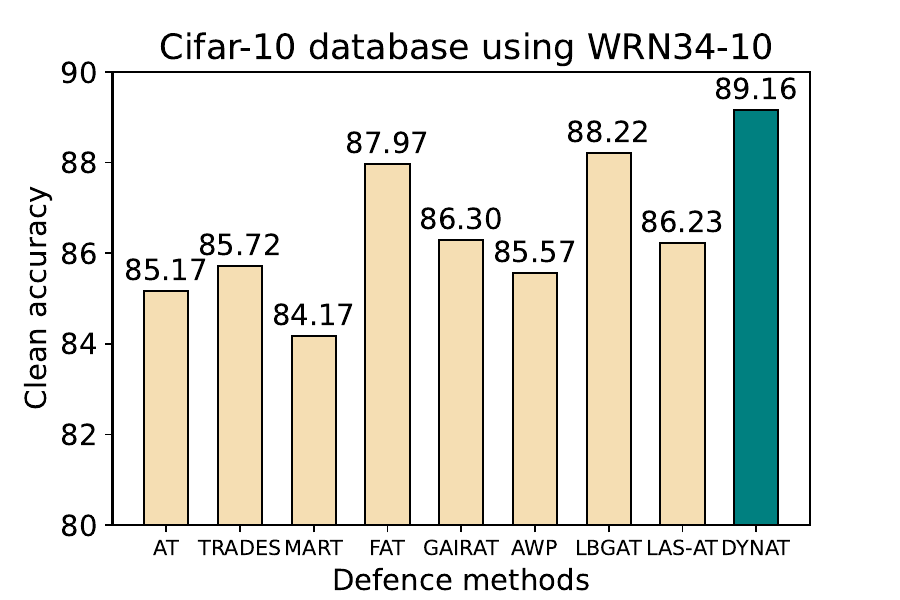}
        \caption{CIFAR-10}
        \label{fig:sat0}
    \end{subfigure} 
\begin{subfigure}[t]{0.47\linewidth}
        \centering
        \includegraphics[width=1.1\linewidth]{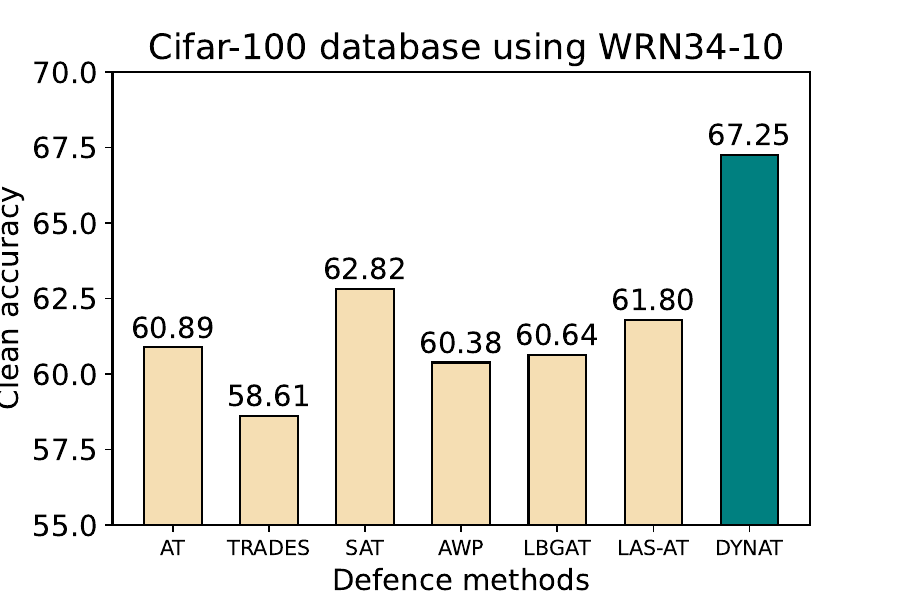}
        \caption{CIFAR-100}
        \label{fig:las-at0}
    \end{subfigure} 
  \caption{Performance of DYNAT on CIFAR-10 and CIFAR-100. (a) Comparison of DYNAT  with the other defense methods, such as AT, TRADES, MART, etc., using the WRN34-10 network on the CIFAR-10 dataset.  (b) Comparison of DYNAT  with the other defense methods, such as AT, TRADES, SAT, etc., using the WRN34-10 network on the CIFAR-100 dataset. In each plot, the y-axis represents the clean accuracy.
  }
\label{fig:wrn34_arch_clean}
\end{figure}

We examine the performance of DYNAT in compassion with other methods on clean accuracy to understand the effect of dynamic labels. The results of this experiment show the dominance of clean accuracy of DYNAT methods (see Fig.~\ref{fig:wrn34_arch_clean}).

\subsection{Comparisons with defense method GAIRAT}
\begin{figure}[t]
\centering
\includegraphics[width=0.5\linewidth]{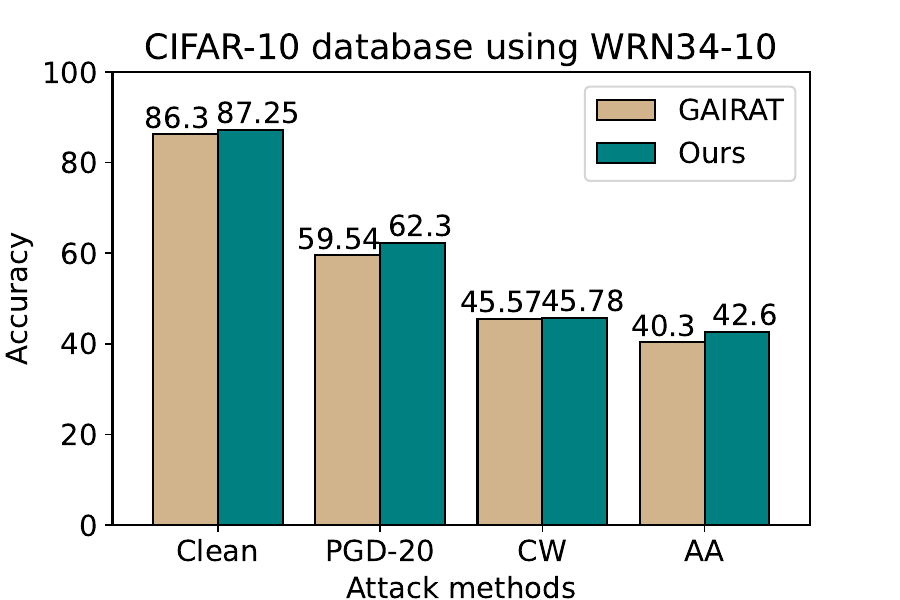}
\caption{Performance of WRN34-10 target model on CIFAR-10 dataset. Given the guide model (ResNet18), the target models (WRN-34-10) were obtained using the GAIRAT method (shown in light-colored bars) and a combination of our method with  GAIRAT (shown in dark colors).}
\label{fig:GAIRAT}
\end{figure}

From the CIFAR-10 dataset results in Table~\ref{tb:cifar10_wrn34}, we noted that the performance of GAIRAT is best on PGD attacks compared to our methods and other methods. Therefore, we experiment with a hybrid method combining DYNAT with GAIRAT to asses if the robustness of the attacks significantly improves. In this combination, we use ResNet18 as the guiding model for DYNAT to provide dynamic labels to WideResNet34-10, which was used as a target model that follows GAIRAT training processes (including all hyperparameter settings). We adopt weight decay of 5 × 10$-4$, label smoothing 0.2, and 70 epochs burn-in period. The results of this experiment are shown in Fig.~\ref{fig:GAIRAT} and Table~\ref{tb:GAIRAT} where it can be observed that our hybrid method outperforms the standalone GAIRAT on all attacks for the CIFAR10 dataset. 
We also performed a separate set of experiments in which we swapped the guiding and target model types. This was to ensure that our method is general and shows distillation from any size of the models. That is, we set ResNet18 as the target model and WRN34-10 as the guiding model. Indeed, the performance of this combined method of defense outperforms the standalone GAIRAT method (see Table~\ref{tb:GAIRAT} row with WRN34-10 as the target network).


\begin{table}[h!]
\setlength{\tabcolsep}{5pt}
\centering
\caption{Test robustness accuracy (\%)  on the CIFAR-10 dataset using WRN34-10. Inner indicates inner optimization as in Sec.~\ref{sec:inner_opt}, and \textbf{bold} indicates the best result.}
\label{tb:GAIRAT}
\begin{tabular}{cc|cccccc}
\toprule
Target Network & Method           & Clean           & PGD-20                & C\&W           & AA             \\ \midrule 

\multirow{2}{*}{WRN34-10~\quad} & GAIRAT \cite{zhang2020geometry}          & 86.30        &          59.54     &        45.57      &          40.30          \\ 

 & DYNAT-Inner-GAIRAT (\textbf{ours})~   & \textbf{87.25}          &       \textbf{60.20}       &       \textbf{45.78}       & \textbf{42.60}                  \\ 
 \midrule 

\multirow{2}{*}{ResNet-18~\quad} & GAIRAT \cite{zhang2020geometry}          & 82.27        &          60.07     &        37.82      &          35.70          \\ 
& DYNAT-Inner-GAIRAT (\textbf{ours})~ & \textbf{82.32}          &       \textbf{62.30}       &       \textbf{38.73}       & \textbf{37.90}                  \\ 
\bottomrule
\end{tabular}
\end{table}

\subsection{Comparison with defense method MAIL} 
In addition to GAIRAT, we notice that MAIL~\cite{liu2021probabilistic} is an improvement to GAIRAT. In comparison to GAIRAT, MAIL performs a noticeable improvement in robustness accuracy under C\&W and AA attacks. Therefore, we conducted a comparative analysis of MAIL with a combination of dynamic and MAIL defense methods to better reflect the results that follow the hyperparameter setting described in~\cite{liu2021probabilistic}.

From the performance of MAIL with WRN34-10 as the guiding model (as shown in Table~\ref{tb:MAIL}), it is evident that dynamic labels significantly enhance the model's robustness against C\&W and AA attacks. Specifically, the accuracy under the AA attack has increased by 4.78 \%, and under the C$\&$W attack, the accuracy has risen by 3.69\%. Furthermore, as mentioned in Sec.~\ref{sec:inner_opt}, our inner optimization contributes to enhancing the robustness, achieving a robustness accuracy of 49.90\% under the AA attack.

\begin{table}[h!]
\centering
\caption{Test robustness (\%)  on the CIFAR-10 database using ResNet-18 as target model. Inner indicates inner optimization, and \textbf{bold} and \underline{underlined} number indicates 1st and 2nd best results.}
\label{tb:MAIL}
\begin{tabular}{ll|cccccc}
\toprule
& Method           & Clean           & PGD-100     & APGD           & C\&W           & AA             \\ \midrule 

\multirow{8}{*}{Others~} & AT ~\cite{rice2020overfitting}      & 84.86       &          48.91     &       47.70      &          51.61    &      44.90         \\

& TRADES ~\cite{zhang2019theoretically}       & 84.00      &          52.66     &       52.37      &          52.30    &      48.10         \\

& MART ~\cite{wang2019improving}      & 82.28      &          53.50     &       52.73      &          51.59    &      48.40         \\

& FAT  ~\cite{zhang2020attacks}       & \textbf{87.97}      &          46.78     &       46.68      &          49.92    &      43.90          \\

& AWP  ~\cite{wu2020adversarial}         & \underline{85.17}       &          52.63     &        50.40      &          51.39    &      47.00          \\

& GAIRAT \cite{zhang2020geometry}        & 83.22       &           \underline{54.81}     &        50.95      &          39.86    &      33.35          \\

& MAIL \cite{liu2021probabilistic}         & 84.52       &           \textbf{55.25}     &        \textbf{53.20}      &          48.88    &      44.22          \\ 
& MAIL (MSE)        &   82.99         &         51.88         &  51.10        &      \underline{52.31}     &     48.90 \\
\midrule 
\multirow{2}{*}{Our Experiments~} 
& DYNAT-AT-MAIL  & 83.71     &            52.49    &        52.40       &        52.13            &  \underline{49.00}   \\ 
& DYNAT-MAIL–Inner   & 82.30        &      53.02           & \underline{52.80}           &     \textbf{52.57}             &   \textbf{49.90} \\
\bottomrule
\end{tabular}
\end{table}

\section{Conclusion}
We propose a novel dynamic adversarial training (DYNAT) algorithm for deep learning security. We discover the limitations of adversarial distillation and adversarial training methods and introduce a dynamic label generation method. We found that our method progressively guides a target model to become robust, avoiding overfitting challenges by using ``\textit{weak to strong}'' label generation. Our approach offers significant robustness to various state-of-the-art attacks. We show that our target model does not necessarily need to inherit the robustness from a large guiding model as any size of guiding model would generate a high-performing dynamic label.  


\bibliographystyle{splncs04}
\bibliography{ref}
\end{document}